\documentclass[sigconf]{acmart}

\AtBeginDocument{%
  }

\copyrightyear{2026}
\acmYear{2026}
\setcopyright{cc}
\setcctype{by}
\acmConference[CIKM '26]
  {Proceedings of the 35th ACM International Conference on Information and Knowledge Management}
  {November 7--11, 2026}
  {Rome, Italy.}
\acmBooktitle{Proceedings of the 35th ACM International Conference on Information and Knowledge Management (CIKM '26), November 7--11, 2026, Rome, Italy}
\acmISBN{979-8-4007-2539-5/2026/11}
\acmDOI{10.1145/3799682.3839976}
\usepackage{colortbl}
\usepackage{siunitx}
\definecolor{GuidelineGreen}{RGB}{34,139,34}

\begin{document}

\title[Scaling Creative Writing Beyond Story-Centric Data]{
Scaling Creative Writing Beyond Story-Centric Data\\with Attribute-Guided Genre Expansion}

\author{Hwan Chang}
\orcid{0009-0009-3954-1975}
\authornote{Work done during internship at LG AI Research.}
\affiliation{%
  \institution{Chung-Ang University}
  \city{Seoul}
  \country{Republic of Korea}}
\email{hwanchang16@gmail.com}

\author{Yongil Kim}
\affiliation{%
  \institution{LG AI Research}
  \city{Seoul}
  \country{Republic of Korea}}
\email{yong-il.kim@lgresearch.ai}

\author{Heuiyeen Yeen}
\affiliation{%
  \institution{LG AI Research}
  \city{Seoul}
  \country{Republic of Korea}}
\email{heuiyeen214@lgresearch.ai}

\author{Yireun Kim}
\affiliation{%
  \institution{LG AI Research}
  \city{Seoul}
  \country{Republic of Korea}}
\email{yireun.kim@lgresearch.ai}

\author{Jinsik Lee}
\affiliation{%
  \institution{LG AI Research}
  \city{Seoul}
  \country{Republic of Korea}}
\email{jinsik.lee@lgresearch.ai}

\author{Hwanhee Lee}
\orcid{0000-0002-9367-9811}
\authornote{Corresponding author.}
\affiliation{%
  \institution{Chung-Ang University}
  \city{Seoul}
  \country{Republic of Korea}}
\email{hwanheelee@cau.ac.kr}

\renewcommand{\shortauthors}{Hwan Chang et al.}

\begin{abstract}
High-quality creative writing data for language models remains dominated by story-centric data, limiting models' ability to follow the structural and functional conventions of diverse creative formats. We propose an \textit{attribute-guided genre expansion} framework for scaling creative writing data beyond story generation. 
By separating thematic breadth from genre-form control, our framework leverages human-authored story prompts as diverse creative seeds, while utilizing manually curated genre attributes to enforce distinct structural, stylistic, and formatting conventions.
We combine these to prompt strong models for genre-faithful query--response pairs, which are then quality-filtered. Applying this framework, we construct the \textit{Multi-Genre Collection}, a 50K-example corpus spanning 13 creative genres, including story, lyrics, game design, and other creative formats. 
Experiments demonstrate that models fine-tuned on our data consistently surpass not only base models and writing-specialized baselines, but also models trained on existing writing corpora.
Genre-count ablations further indicate that genre expansion is a key driver of robust creative writing capability.
\end{abstract}




\maketitle
\begin{figure}
  \centering
  \includegraphics[width=\columnwidth]{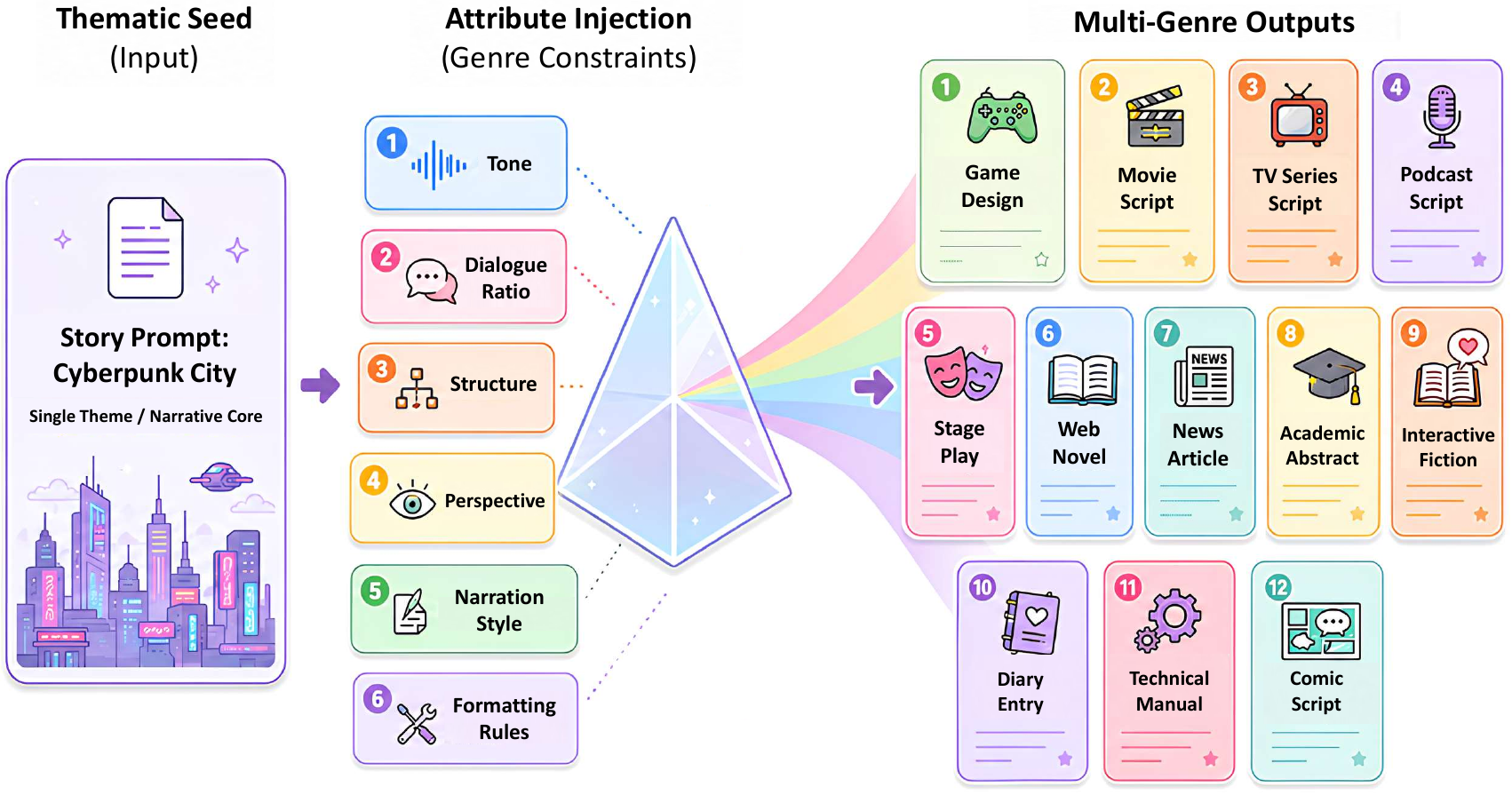}
  \vspace{-18pt}
  \caption{Overview of our attribute-guided multi-genre expansion framework.
A thematic seed is transformed into diverse outputs by injecting curated genre attributes.}
  \Description{Overview diagram showing how thematic seeds are transformed into genre-faithful outputs through curated genre attributes.}
  \label{fig:overview_figure}
  \vspace{-4mm}
\end{figure}

\begin{figure*}[!h]
    \centering
    \includegraphics[width=\linewidth]{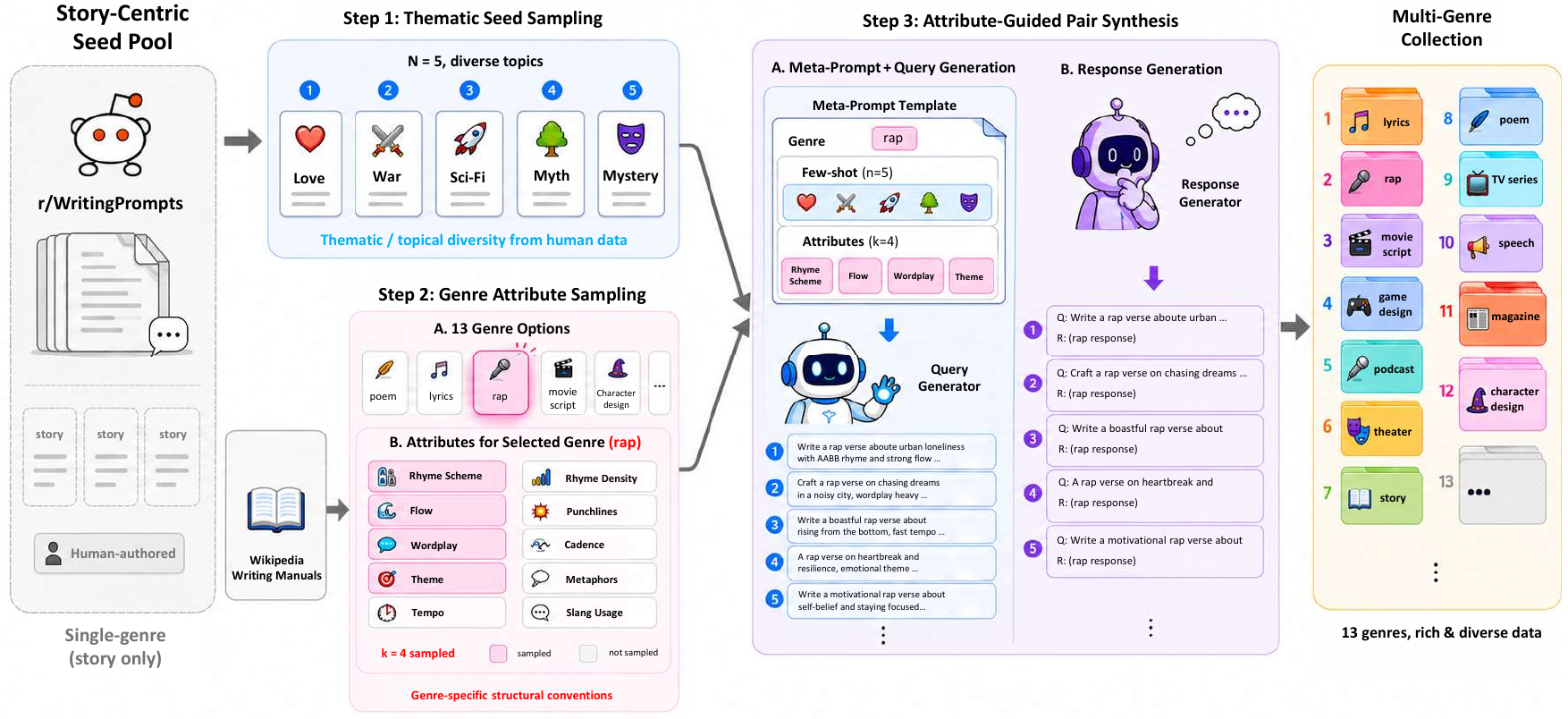}
    \vspace{-18pt}
    \caption{Attribute-guided genre expansion pipeline. We sample few-shot examples from existing story prompts to maintain thematic diversity, randomly sample subset genre-specific attributes to inject structural constraints, and generate multi-genre queries by combining task instructions, few-shot demonstrations, target genre, and sampled attributes.}
    \Description{Pipeline diagram illustrating few-shot sampling, genre-attribute injection, and query generation for attribute-guided multi-genre data construction.}
    \vspace{-3mm}
    \label{fig:data_construction_pipeline}
\end{figure*}

\section{Introduction}
Creative writing is one of the most prominent use cases of large language models (LLMs)~\citep{qwen3technicalreport,exaone-3.5}, accounting for a substantial portion of real-world interactions~\citep{anthropic2026aeiv4,chatterji2025people}. Existing public resources, nonetheless, remain heavily story-centric, largely treating creative writing as narrative generation~\citep{fan2018hierarchical,mostafazadeh-etal-2016-corpus}. Yet real-world creative writing extends far beyond stories~\citep{wu2025writingbench}: users ask models to write rap verses, game design documents, and other creative formats. These genres differ not only in topic or style, but also in their structural and functional conventions—rap demands rhyme scheme and flow, movie scripts rely on narrative arc and dialogue, game design documents require mechanics and player interaction. 
A model trained primarily on story-centric data may thus learn rich narrative content while failing to follow the genre-specific structural and functional constraints expected in practical creative writing tasks.

However, mere scaling of story-centric data fails to address these structural gaps, as it does not systematically encode the formal constraints of non-story genres.
Existing synthetic instruction-generation methods offer a natural starting point by bootstrapping seed tasks, evolving instructions, or rewriting problems at scale~\citep{wang2023selfinstruct,xu2023wizardlm}.
Yet, many successful applications have targeted general instruction following or verifiable domains such as code and mathematics, where difficulty, correctness, or solution validity can be approximated through seed examples, executable constraints, final answers, or problem transformations~\citep{luo2023wizardcoder,wei2023magicoder,yu2023metamath,toshniwal2024openmathinstruct}.
Creative writing lacks such compact validity signals. Directly applying generic synthesis pipelines risks producing prompts that are topically diverse but formally under-specified, because genre fidelity depends on human-recognizable conventions such as lyric flow, screenplay formatting, and game mechanics. These limitations suggest that creative writing data generation requires human guidance not merely for quality filtering, but for defining the genre-form constraints that generation should follow.

To bridge this gap and effectively scale creative writing capabilities, we identify three key requirements. First, \emph{systematic genre coverage} is necessary to represent the range of creative formats users actually request. Second, \emph{genre-form fidelity} is needed because each genre imposes distinct structural, stylistic, functional, and formatting constraints. Third, \emph{specification diversity} is necessary because real writing instructions naturally range from open-ended requests to highly constrained prompts. These requirements call for a controlled way to expand story-centric creative prompts into genre-faithful instruction data.

To address this challenge, we propose \textbf{attribute-guided genre expansion}, a controlled generation framework for scaling creative writing data beyond story-centric sources. Our framework is related to synthetic instruction generation, but differs in its axis of control: rather than evolving prompts primarily by task complexity or problem variation, we separate thematic variation from genre-form control. Human-authored story prompts provide diverse creative seeds, while manually curated attributes specify the conventions of each target genre. For each genre, we draw seeds from a curated \texttt{r/WritingPrompts} pool and sample a subset of its attributes, then combine them with the target genre to generate genre-faithful writing queries. The queries are paired with responses from strong LLMs and quality-filtered to yield reliable query--response data. Applying this framework, we construct the \textit{Multi-Genre Collection}, a 50K-instance corpus spanning 13 creative writing genres.

Our experiments show that the resulting data improves creative writing capability across multiple dimensions. 
Models fine-tuned on the \textit{Multi-Genre Collection} consistently outperform not only their base models and the writing-specialized LongWriter-glm4-9B~\citep{bai2025longwriter} baseline, but also competitor models trained on existing writing corpora~\citep{wang2026reverseengineered} across all popular benchmarks.
We further find that increasing genre coverage improves output novelty: as the number of training genres increases from story-only to all 13 genres, novelty metrics~\citep{zhang2025noveltybench} improve consistently.
Finally, evaluations with independent LLM judges and humans confirm these gains reflect genuine quality improvements rather than evaluator bias.

\section{Attribute-Guided Genre Expansion}
\label{sec:method}
\label{sec:data}

\begin{figure}[!t]
\centering
\begingroup
\setlength{\fboxrule}{0.3pt}
\setlength{\fboxsep}{5pt}
\fbox{%
\begin{minipage}{0.90\linewidth}
\footnotesize

You are a \textcolor{red}{\{genre\}} query generator. \\
\textcolor{red}{\{task\_instruction\}} \\
\textcolor{GuidelineGreen}{(Guideline: Create distinct scenarios.)} \\
\vspace{4pt}
\noindent When crafting each query, incorporate these specific dimensions: \\
\textcolor{red}{\{genre\_attribution\}} 

\textcolor{GuidelineGreen}{(0 $\sim$ len(genre\_attribute) genre-specific attributes sampled)} \\

\vspace{4pt}

\noindent Draw thematic inspiration from the examples below, but reimagine them with new perspectives suitable for \textcolor{red}{\{genre\}}.

\vspace{4pt}

\textcolor{red}{\{few\_shot\}} \\
\textcolor{GuidelineGreen}{(Sampled from existing Story Generation Query Pool for thematic variety)}

\end{minipage}%
}
\endgroup
\vspace{-12pt}
\caption{Abstract Structured Meta-Prompt Template.}
\Description{A meta-prompt template that combines a target genre, task instructions, sampled genre attributes, and few-shot story examples.}
\vspace{-3mm}
\label{fig:query_generation_template}

\end{figure}

We formulate the construction of the Multi-Genre Collection as an attribute-guided genre expansion process for scaling creative writing beyond story-centric data. Rather than directly collecting prompts for each creative genre, our framework starts from a story-centric human-authored seed pool and generates genre-faithful creative writing query--response pairs through the three-stage process illustrated in Figure~\ref{fig:data_construction_pipeline}. Instantiating this process yields an English corpus spanning 13 creative genres.

Formally, let $(q^{g}, y)$ denote a query--response pair for genre $g$ (e.g., $q^{\texttt{rap}} = $ \textit{``Write a rap verse about urban loneliness with an AABB rhyme scheme''}). Prior writing datasets are heavily concentrated on $g = \texttt{story}$; our goal is to expand coverage beyond story writing by generating new queries for each target genre through attribute-guided genre expansion. 

Let $X = \{(q_i^{\texttt{story}}, y_i)\}_{i=1}^{n}$ denote few-shot examples sampled from the seed dataset, and let $A \subseteq A_g$ denote a subset of genre-specific attributes for genre $g$. We construct a meta-prompt via a template $\mathcal{T}$ (Figure~\ref{fig:query_generation_template}) and generate queries through an LLM:
\begin{equation}
  Q = \mathrm{LLM}\bigl(\mathcal{T}(g, X, A)\bigr).
  \label{eq:query_gen}
\end{equation}
The meta-prompt $\mathcal{T}(g, X, A)$ unifies thematic seeds from human-authored data with genre-specific structural constraints. In this way, the pipeline separates two sources of variation: the seed examples provide topical and stylistic breadth, while the sampled genre attributes control the structural conventions of the target genre.

\paragraph{\textbf{Source Dataset Selection.}}
We choose Reddit's community forum \texttt{r/WritingPrompts} as our seed dataset for $(q^{\texttt{story}}, y)$ pairs, as it provides human-authored prompts with high topical diversity and naturally varying specificity. To ensure quality, we apply two filters using GPT-5-mini~\citep{singh2025openai}: (1)~\textit{safety filtering}, which removes 686 instances containing harmful or inappropriate content, and (2)~\textit{irrelevant content removal}, which cleans 861 instances of off-topic fragments (e.g., meta-commentary or subreddit boilerplate). This curated seed pool serves as the source of thematic diversity for subsequent genre transfer.

\paragraph{\textbf{Step 1:  Thematic Seed Sampling}}
For each template instantiation, we randomly sample $n{=}5$ query--response pairs from the curated seed dataset to serve as thematic seeds. These seeds transfer the topical and stylistic breadth of human-authored story prompts to non-story creative formats, supplying the thematic variety that drives the generated queries.

\paragraph{\textbf{Step 2: Genre Attribute Sampling.}}
Since different genres are governed by distinct structural conventions, we define \textit{genre attributes} as the key dimensions along which queries within a genre can meaningfully vary---for example, \emph{rhyme scheme} and \emph{flow} for rap, or \emph{narrative arc} and \emph{character development} for TV series. To construct these attributes, we collect authoritative genre definitions from encyclopedic references (e.g., Wikipedia) and creative writing manuals, then prompt GPT-5~\citep{singh2025openai} to extract structured attribute lists from each definition. The resulting attributes are all manually reviewed and refined to verify that each reflects a genuine structural convention of the genre, to merge overlapping attributes, to remove overly generic ones, and to add any salient dimensions missed by the automatic extraction---ensuring coverage and genre fidelity and yielding a curated set $A_g$ of 5--15 attributes per genre $g$. To reflect natural variation in instruction specificity, we sample $k \sim \mathrm{Uniform}(0, |A_g|)$ and then randomly select a subset $A \subset A_g$ with $|A| = k$. When $k = 0$, the query remains open-ended; as $k$ increases, the query becomes progressively more constrained. Thus, attribute sampling acts as the pipeline's control mechanism for varying instruction specificity while preserving genre-specific structure.

% \begin{table}
%     \centering
%     \caption{Performance across out-of-distribution generalization benchmarks, and in-distribution genre coverage.}
%     \label{tab:performance}
%     \small
%     \renewcommand{\arraystretch}{1.1}
%     \begin{tabular}{l r@{\,}l r@{\,}l r@{\,}l}
%         \toprule
%         & \multicolumn{4}{c}{\textbf{OOD (Generalization)}}
%             & \multicolumn{2}{c}{\textbf{ID (Coverage)}} \\
%         \cmidrule(lr){2-5}\cmidrule(lr){6-7}
%         \textbf{Models} 
%             & \multicolumn{2}{c}{\textbf{Arena Hard}} 
%             & \multicolumn{2}{c}{\textbf{WritingBench}} 
%             & \multicolumn{2}{c}{\textbf{Multi-Genre}} \\
%         \midrule
%         LongWriter   & 2.3  && 47.2 && 49.5 & \\
%         \midrule
%         Llama        & 2.9  && 43.7 && 44.4 & \\
%         Llama + SFT  & 33.0 & {\tiny\textcolor{red}{(+30.1)}} & 60.6 & {\tiny\textcolor{red}{(+16.9)}} & 68.4 & {\tiny\textcolor{red}{(+24.0)}} \\
%         EXAONE       & 10.4 && 52.7 && 64.9 & \\
%         EXAONE + SFT & 22.9 & {\tiny\textcolor{red}{(+12.5)}} & 59.4 & {\tiny\textcolor{red}{(+6.7)}}  & 66.2 & {\tiny\textcolor{red}{(+1.3)}}  \\
%         Qwen3        & 8.0  && 56.1 && 67.7 & \\
%         Qwen3 + SFT  & \textbf{34.2} & {\tiny\textcolor{red}{(+26.2)}} & \textbf{63.6} & {\tiny\textcolor{red}{(+7.5)}} & \textbf{69.3} & {\tiny\textcolor{red}{(+1.6)}} \\
%         \bottomrule
%     \end{tabular}
% \end{table}
\paragraph{\textbf{Step 3: Attribute-Guided Pair Synthesis.}}
Given the few-shot examples $X$ and sampled attributes $A$, we prompt GPT-5-mini~\citep{singh2025openai} as the query generator $\mathrm{LLM}(\cdot)$ in Eq.~\eqref{eq:query_gen} to produce five queries per template instantiation. To further promote diversity and mitigate model collapse, we employ verbalized sampling~\citep{zhang2025verbalizedsamplingmitigatemode}---explicitly instructing the model to produce outputs that vary in topic, tone, and structure. 
We then generate responses using Qwen3-235B-A22B-Thinking~\citep{qwen3technicalreport}. To filter low-quality responses, we employ an independent LLM-as-a-judge~\citep{zheng2023judging} using Qwen3-30B-A3B-Instruct~\citep{qwen3technicalreport} to evaluate response quality and exclude pairs whose scores fall below two standard deviations from the mean.

\begin{figure}[!h]
  \centering
  \includegraphics[width=0.8\columnwidth]{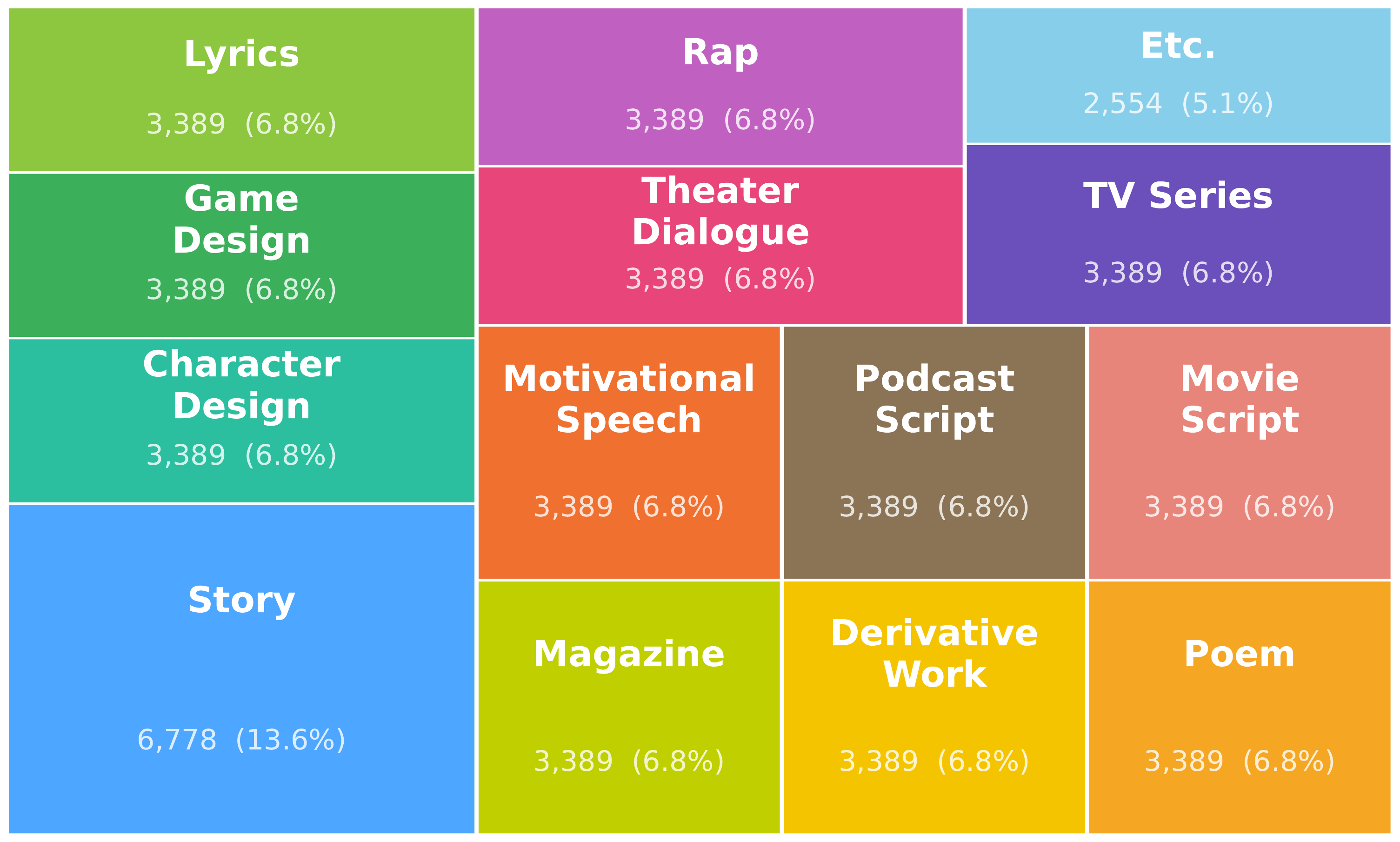}
  \vspace{-4pt}
  \caption{Genre distribution in the Multi-Genre Collection.}
  \Description{Distribution chart showing that the Multi-Genre Collection contains balanced coverage across 13 creative writing genres.}
  \vspace{-3mm}
  \label{fig:statistics}
\end{figure}

\begin{figure}[!h]
  \centering
  \includegraphics[width=0.9\columnwidth]{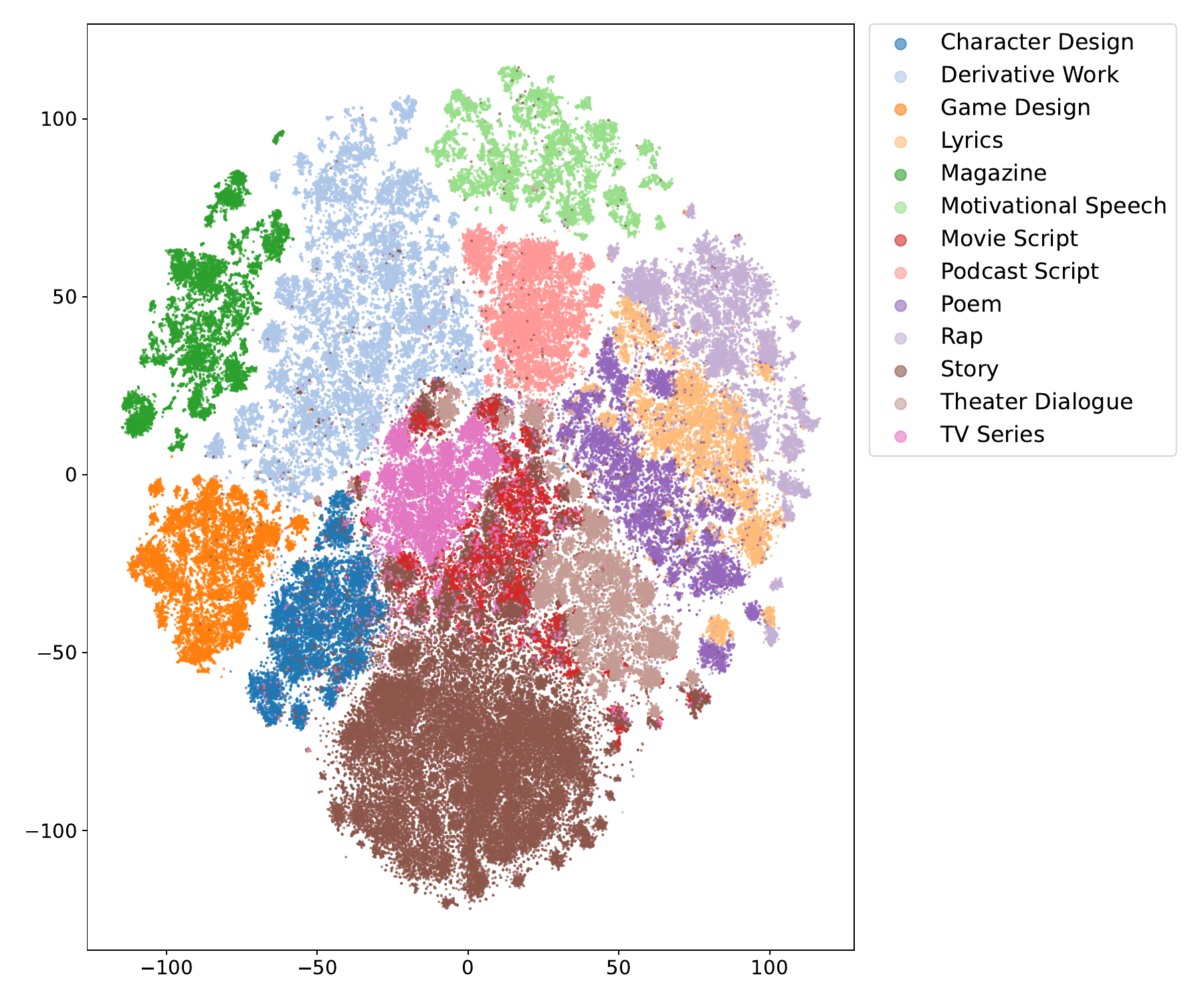}
  \vspace{-9pt}
  \caption{t-SNE visualization of the Multi-Genre Collection.}
  \Description{t-SNE scatter plot showing semantic clusters for the 13 genres in the Multi-Genre Collection.}
  \vspace{-3mm}
  \label{fig:tsne_embedding}
\end{figure}

\paragraph{\textbf{Synthesis Outcomes and Diversity Analysis.}}
Instantiating the above pipeline yields the Multi-Genre Collection, a 50K-instance creative writing corpus with balanced coverage across the target genres (Figure~\ref{fig:statistics}). The 'Etc.' category captures long-tail formats (e.g., diary entries, comic scripts, interactive fiction) outside our 13 primary genres. To verify semantic distinctiveness, we embed all queries using Qwen3-Embedding-0.6B~\citep{qwen3embedding} and project them into 2D via t-SNE (Figure~\ref{fig:tsne_embedding}), revealing clear genre-aligned clusters with minimal overlap. This indicates that the attribute-guided synthesis pipeline produces genre-distinct instruction distributions rather than merely paraphrasing story-centric prompts.

\section{Experiments}

\subsection{Experimental Setup}
\paragraph{\textbf{Models.}}
We evaluate three base models: Llama-3.1-8B-Instruct (Llama)~\citep{grattafiori2024llama}, EXAONE-3.5-7.8B-Instruct (EXAONE)~\citep{exaone-3.5}, and Qwen3-8B (Qwen)~\citep{qwen3technicalreport}. All models are fine-tuned via Supervised Fine-Tuning (SFT) using LlamaFactory~\citep{zheng2024llamafactory} with LoRA~\citep{hu2022lora} adapters of rank 128 applied to all transformer layers, preserving pre-trained knowledge while ensuring computational efficiency. We train for 2 epochs with a cutoff length of 4096 tokens, learning rate $1.0 \times 10^{-5}$, warmup ratio 0.1, and global batch size 128. As a writing-specialized baseline, we additionally include LongWriter-glm4-9B~\citep{bai2025longwriter}.

\paragraph{\textbf{Benchmarks.}}
We evaluate models across three benchmarks: (1)~\textit{Arena Hard (Creative Writing)}~\citep{li2025arenahard} v2.0, filtered to English samples, judged by GPT-4.1 against reference answers following the original evaluation protocol; (2)\textit{WritingBench}~\citep{wu2025writingbench}, filtered to English samples in the creative writing domains (\textit{Advertising \& Marketing} and \textit{Literature \& Arts}), judged by GPT-5-mini following the original evaluation protocol; and (3)~the held-out test set from our \textit{Multi-Genre Collection} (Multi-Genre), spanning all 13 genres with 50 instances per genre (650 total), where Qwen3-235B-A22B-Thinking outputs serve as references and GPT-4.1 rates each model output on a 1--10 scale for quality, creativity, and genre adherence.

\subsection{Experimental Results}
\label{subsec:results}

\begin{table}[!t]
    \centering
    \caption{Performance across out-of-distribution generalization benchmarks, and in-distribution genre coverage.}
    \label{tab:performance}
    \small
    \renewcommand{\arraystretch}{1.1}
    \begin{tabular}{l
        S[table-format=2.1, table-space-text-post={\,\tiny(+00.0)}]
        S[table-format=2.1, table-space-text-post={\,\tiny(+00.0)}]
        S[table-format=2.1, table-space-text-post={\,\tiny(+00.0)}]}
        \toprule
        & \multicolumn{2}{c}{\textbf{OOD (Generalization)}}
            & \multicolumn{1}{c}{\textbf{ID (Coverage)}} \\
        \cmidrule(lr){2-3}\cmidrule(lr){4-4}
        \textbf{Models}
            & \multicolumn{1}{c}{\textbf{Arena Hard}}
            & \multicolumn{1}{c}{\textbf{WritingBench}}
            & \multicolumn{1}{c}{\textbf{Multi-Genre}} \\
        \midrule
        LongWriter   & 2.3  & 47.2 & 49.5 \\
        \midrule
        Llama        & 2.9  & 43.7 & 44.4 \\
        \rowcolor{gray!15}
        Llama + SFT  & {33.0\,{\tiny\textcolor{red}{(+30.1)}}} & {60.6\,{\tiny\textcolor{red}{(+16.9)}}} & {68.4\,{\tiny\textcolor{red}{(+24.0)}}} \\
        EXAONE       & 10.4 & 52.7 & 64.9 \\
        \rowcolor{gray!15}
        EXAONE + SFT & {22.9\,{\tiny\textcolor{red}{(+12.5)}}} & {59.4\,{\tiny\textcolor{red}{(+6.7)}}}  & {66.2\,{\tiny\textcolor{red}{(+1.3)}}}  \\
        Qwen3        & 8.0  & 56.1 & 67.7 \\
        \rowcolor{gray!15}
        Qwen3 + SFT  & {\textbf{34.2}\,{\tiny\textcolor{red}{(+26.2)}}} & {\textbf{63.6}\,{\tiny\textcolor{red}{(+7.5)}}} & {\textbf{69.3}\,{\tiny\textcolor{red}{(+1.6)}}} \\
        \bottomrule
    \end{tabular}
\end{table}
\begin{table}
    \centering
    \vspace{-1.5mm}
    \caption{Comparison with existing writing data using 2K sampled subsets, with Qwen3-8B fine-tuned on each corpus.}
    \vspace{-2.0mm}
    \label{tab:comparison}
    \small
    \renewcommand{\arraystretch}{1.1}
    \begin{tabular}{lccc}
        \toprule
        & \multicolumn{2}{c}{\textbf{OOD (Generalization)}}
            & \multicolumn{1}{c}{\textbf{ID (Coverage)}} \\
        \cmidrule(lr){2-3}\cmidrule(lr){4-4}
        \textbf{Data} & \textbf{Arena Hard} & \textbf{WritingBench} & \textbf{Multi-Genre} \\
        \midrule
        DeepWriting & 4.9  & 55.5 & 65.9 \\
        LongWriter  & 7.0  & 56.4 & 69.6 \\
        \rowcolor{gray!15}
        Multi-Genre & \textbf{7.5} & \textbf{59.1} & \textbf{71.4} \\
        \bottomrule
    \end{tabular}
    \vspace{-1mm}
\end{table}
\paragraph{\textbf{Overall Performance.}}
Table~\ref{tab:performance} shows that all three base models improve substantially after SFT on our Multi-Genre Collection, with consistent gains on out-of-distribution benchmarks (largest for Qwen3-8B) and balanced in-distribution coverage across all 13 genres. Notably, all fine-tuned models outperform the writing-specialized LongWriter-glm4-9B~\citep{bai2025longwriter} baseline, confirming that diverse, genre-specific data is more effective than narrow writing specialization.

% \paragraph{\textbf{Comparison with Existing Writing Datasets.}}
% As shown in Table~\ref{tab:comparison}, Qwen3-8B fine-tuned on Multi-Genre Collection substantially outperforms models trained on DeepWriting-20k~\citep{wang2026reverseengineered} and LongWriter-6k~\citep{bai2025longwriter} across all benchmarks, with particularly large gains on Arena Hard, demonstrating that our attribute-guided pipeline produces higher-quality, more transferable training data.
\paragraph{\textbf{Comparison with Existing Writing Datasets.}}
Since the Multi-Genre Collection is considerably larger, we randomly sample 2K examples from each dataset for SFT to ensure a fair comparison. As shown in Table~\ref{tab:comparison}, Qwen3-8B fine-tuned on Multi-Genre Collection substantially outperforms models trained on DeepWriting-20k~\citep{wang2026reverseengineered} and LongWriter-6k~\citep{bai2025longwriter} across all benchmarks, with particularly large gains on Arena Hard, demonstrating that our attribute-guided pipeline produces higher-quality, more transferable training data.

\begin{table}[!t]
    \centering
    %\vspace{-1.5mm}
    \caption{Effect of genre count on creative output novelty. More genres yield more varied, non-redundant outputs.}
    \vspace{-1.5mm}
    \label{tab:novelty}
    \small
    \renewcommand{\arraystretch}{1.1}
    \begin{tabular}{lccccc}
        \toprule
        \textbf{\# Genres} & 0 & 4 & 8 & 12 & All \\
        \midrule
        \textbf{Novelty} ($\uparrow$) & 3.87 & 3.93 & 4.26 & 4.56 & \textbf{4.81} \\
        \bottomrule
    \end{tabular}
\end{table}
\begin{table}
    \centering
    \vspace{-0.5mm}
    \caption{WritingBench evaluation under independent judges.}
    \vspace{-2.0mm}
    \label{tab:judge_bias}
    \small
    \renewcommand{\arraystretch}{1.1}
    \begin{tabular}{lccc}
        \toprule
        \textbf{Judge} & \textbf{GPT-5-mini} & \textbf{DeepSeek v3.2} & \textbf{Gemini 3 Flash} \\
        \midrule
        Llama        & 43.7 & 36.6 & 41.7 \\
        \rowcolor{gray!15}
        Llama + SFT  & 60.6 & 63.1 & 70.5 \\
        Qwen         & 56.1 & 51.5 & 58.5 \\
        \rowcolor{gray!15}
        Qwen + SFT   & \textbf{63.6} & \textbf{65.9} & \textbf{71.5} \\
        \bottomrule
    \end{tabular}
\end{table}
\begin{table}
    \centering
    \caption{Human evaluation on WritingBench.}
    \vspace{-2.0mm}
    \label{tab:human_eval}
    \small
    \renewcommand{\arraystretch}{1.1}
    \begin{tabular}{lccc}
        \toprule
        \textbf{Data} & \textbf{Multi-Genre} & \textbf{LongWriter} & \textbf{DeepWriting} \\
        \midrule
        Human Eval & \textbf{59.3} & 57.7 & 56.9 \\
        \bottomrule
    \end{tabular}
\end{table}
\begin{figure*}[!t]
\centering
\definecolor{QualitativeRed}{RGB}{164,31,54}
\setlength{\fboxsep}{4pt}
\fbox{%
\begin{minipage}{0.97\textwidth}
\scriptsize
\textbf{Prompt:} Write detailed, story-driven symphonic metal deathcore lyrics about a young wizard corrupted by growing power, formatted for Suno.\\[-1mm]
\rule{\linewidth}{0.3pt}
\begin{minipage}[t]{0.48\linewidth}
\colorbox{black!65}{\parbox{\dimexpr\linewidth-2\fboxsep\relax}{\centering\color{white}\textbf{\texttt{Llama} (Base)}}}\\[0.5mm]
\textbf{Verse 1:} In hallowed halls of ancient lore / A young apprentice, I walked before / Masters of the arcane, guiding my hand\ldots\\
\textbf{Chorus:} I rise, a phoenix, from the ashes born / As knowledge grows, so does the darkness sworn\ldots\\
\textbf{Outro:} A legacy of darkness, forever to last / The price of power, a soul forever lost\ldots
\end{minipage}\hfill
\begin{minipage}[t]{0.48\linewidth}
\colorbox{QualitativeRed}{\parbox{\dimexpr\linewidth-2\fboxsep\relax}{\centering\color{white}\textbf{\texttt{Llama + SFT}}}}\\[0.5mm]
\textbf{\textcolor{QualitativeRed}{ASCENDANCE}} \textcolor{QualitativeRed}{(Suno Format)}\\
\textbf{\textcolor{QualitativeRed}{Intro [0:00--0:45]:}} In twilight halls of \textbf{\textcolor{QualitativeRed}{Aethel}}, where shadows learn to speak\ldots\\
\textbf{\textcolor{QualitativeRed}{Verse 1 [0:46--1:30]:}} He mastered the \textbf{\textcolor{QualitativeRed}{Lumen}} spell\ldots{} Wove \textbf{\textcolor{QualitativeRed}{Aero}} currents\ldots{} The \textbf{\textcolor{QualitativeRed}{Verdant}} incantation\ldots\\
\textbf{\textcolor{QualitativeRed}{Bridge [4:36--5:10]:}} He sees \textbf{\textcolor{QualitativeRed}{Elara}}'s smile\ldots{} The child he once protected\ldots
\end{minipage}
\end{minipage}}
\caption{Qualitative comparison on an Arena Hard song-lyrics prompt. \textbf{\textcolor{QualitativeRed}{Dark red}} highlights SFT-distinctive features: a title, named world and entities, and Suno-compatible timestamps.}
\Description{Side-by-side textual comparison of base Llama and Llama plus SFT outputs for a Suno-formatted song-lyrics prompt. Bold dark-red text highlights the SFT output's title, named setting, named character and spells, and section timestamps.}
\label{fig:qualitative}
\vspace{-2mm}
\end{figure*}
\paragraph{\textbf{Impact of Genre Diversity.}}
To examine whether exposure to diverse genres generalizes to broader creative ability, we evaluate outputs using the NoveltyBench Distinct metric~\citep{zhang2025noveltybench}, which partitions generations into semantically and functionally equivalent clusters to estimate output distinctness. As shown in Table~\ref{tab:novelty}, the novelty score rises monotonically as the number of training genres increases from 0 to 13, confirming that diverse creative formats directly enhance the model's capacity to generate varied, non-redundant outputs.

\paragraph{\textbf{Robustness to Judge and Human Evaluation.}}

To address potential self-preference bias from GPT-5-mini serving as both query generator and judge, we re-evaluate on WritingBench using two independent judges (Table~\ref{tab:judge_bias}). All fine-tuned variants strictly outperform their base models across all three evaluators, confirming that gains reflect genuine quality improvements rather than evaluator artifacts. We further validate via human evaluation on 50 prompts randomly sampled from WritingBench, scored following its protocol (Table~\ref{tab:human_eval}), which corroborates these findings.

\paragraph{\textbf{Qualitative Analysis.}}
Figure~\ref{fig:qualitative} further illustrates these gains. Given a prompt requesting story-driven symphonic metal deathcore lyrics formatted for Suno, the base Llama follows a conventional verse--chorus structure but relies on generic imagery and provides neither a title nor named entities. In contrast, Llama + SFT produces a titled composition (\textit{ASCENDANCE}), grounds the narrative in a named setting (\textit{Aethel}), introduces a recurring character (\textit{Elara}) and specific spells (\textit{Lumen}, \textit{Aero}, and \textit{Verdant}), and supplies section-level timestamps. These differences indicate that multi-genre training improves format compliance, world-building specificity, and narrative coherence beyond the aggregate evaluation gains.

% \section{Related Work}
% High-quality datasets have been central to advancing creative writing in language models, though early efforts focused almost exclusively on story generation. WritingPrompts~\citep{fan2018hierarchical} provided diverse prompt--story pairs from Reddit, ROCStories~\citep{mostafazadeh2016corpus} offered five-sentence commonsense stories for narrative understanding, and LitBench~\citep{fein2025litbench} introduced preference labels for evaluating story quality and writing personalization. Beyond stories, prior work has targeted individual genres in isolation: SongComposer~\citep{ding2024songcomposer} for lyric--melody pairs, \citet{mirowski2023co} for screenplay and theatre co-writing, and EssayBench~\citep{gao2025essaybench} for essays. In contrast, the Multi-Genre Collection addresses the field's collective blind spot in a unified framework, spanning 13 diverse English creative genres within a single resource. More critically, we propose a principled attribute-guided genre expansion pipeline that jointly controls genre diversity, thematic breadth, and instruction granularity—dimensions no existing dataset addresses together.
\section{Related Work}
High-quality datasets have been central to advancing creative writing in language models, though early efforts focused almost exclusively on story generation. WritingPrompts~\citep{fan2018hierarchical} provided diverse prompt--story pairs from Reddit, ROCStories~\citep{mostafazadeh-etal-2016-corpus} offered five-sentence commonsense stories for narrative understanding, and LitBench~\citep{fein2025litbench} introduced preference labels for evaluating story quality and writing personalization. Beyond stories, prior work has targeted individual genres in isolation: SongComposer~\citep{ding2024songcomposer} for lyric--melody pairs, and \citet{mirowski2023co} for screenplay and theatre co-writing. In contrast, the Multi-Genre Collection addresses the field's collective blind spot in a unified framework, spanning 13 diverse creative genres within a single resource.

\section{Conclusion}
% We introduce the Multi-Genre Collection, a 50K-instance dataset spanning 13 creative writing genres, built via an attribute-guided pipeline that transfers thematic diversity while enforcing genre-specific constraints. Experiments demonstrate that models fine-tuned on our dataset outperform both base models and writing-specialized baselines, and that increasing genre diversity directly enhances output novelty.
We introduce the Multi-Genre Collection, a 50K-instance, 13-genre dataset. Using our attribute-guided genre expansion, we transfer thematic diversity from human-authored prompts while enforcing genre-specific constraints through manually curated attributes. Our experiments demonstrate three key findings: (1)~models fine-tuned on our dataset substantially outperform both base models and writing-specialized baselines; (2)~our dataset consistently outperforms existing writing datasets; and (3)~increasing genre diversity directly enhances output novelty.

\begin{acks}
This work was supported by the Korea Institute for Advancement of Technology (KIAT) grant funded by the Korea Government (MOTIE) (RS-2025-25458133) and the Institute of Information \& Communications Technology Planning \& Evaluation (IITP) grant funded by the Korea government (MSIT) [RS-2021-II211341, Artificial Intelligence Graduate School Program (Chung-Ang University)].
\end{acks}

% These results establish genre diversity as a critical factor for robust creative writing capabilities in large language models.

% \section*{Limitations}
% While we broaden coverage to 13 genres, creative writing is inherently open-ended, and some niche or emerging formats remain uncovered. Still, our attribute-guided framework is modular and readily extensible to additional genres. 
% The pipeline also relies on LLM-based generation and manually curated attributes, which may introduce subtle biases from seed data or generation models. We mitigate this through filtering, cross-model generation, and diversity-promoting strategies, though minor artifacts may persist.

\section*{GenAI Usage Disclosure}
We write the manuscript ourselves; a general-purpose LLM (ChatGPT) is used solely for refinement of style, clarity, and grammar. It is not used for ideation, claim generation, or experimental design.

As part of the proposed methodology, large language models are used as components of the data construction pipeline---namely, GPT-5-mini for filtering and query generation, Qwen3-235B-A22B-Thinking for response generation, and Qwen3-30B-A3B-Instruct as a quality-scoring LLM-as-a-judge.

% \appendix

% \section{Experimental Details}
% \label{app:experimental_details}
% \paragraph{Training Configuration.}
% All models are fine-tuned using LlamaFactory~\citep{zheng2024llamafactory} with LoRA~\citep{hu2022lora} adapters of rank 128 applied to all transformer layers. We train for 2 epochs at a 4096-token cutoff with a learning rate of $1.0 \times 10^{-5}$, global batch size 128, AdamW with a linear schedule and 0.1 warmup ratio, and gradient checkpointing.

% \paragraph{Benchmark Evaluation.}
% For Arena Hard v2.0 (\texttt{creative\_writing}) we use GPT-4.1 as judge. For WritingBench we evaluate English samples in the \textit{Advertising \& Marketing} and \textit{Literature \& Arts} domains with GPT-5-mini; the independent-judge study (Table~\ref{tab:judge_bias}) reuses this subset under DeepSeek-v3.2 and Gemini-3-Flash. For the Multi-Genre test set we use Qwen3-235B-A22B-Thinking references and GPT-4.1 as judge on a 1--10 scale across quality, creativity, and genre adherence.

% \paragraph{Genre Attributes \& Test Set.}
% Each genre has 5--15 attributes (e.g., \textit{rhyme scheme, flow} for Rap; \textit{verse/chorus form, hook} for Lyrics). We sample $k \sim \mathrm{Uniform}(0, |A_g|)$ attributes per query, spanning open-ended to fully constrained. The held-out test set contains 50 instances per genre (650 total), each pairing a generated query with a Qwen3-235B-A22B-Thinking reference response, enabling balanced cross-genre evaluation.

\bibliographystyle{ACM-Reference-Format}
\bibliography{custom}

\end{document}